\documentclass[sigconf]{acmart}

\AtBeginDocument{%
  }

\copyrightyear{2026}
\acmYear{2026}
\setcopyright{cc}
\setcctype{by-nc-nd}
\acmConference[UbiComp Companion '26]{Companion of the 2026 ACM International Joint Conference on Pervasive and Ubiquitous Computing}{October 11--15, 2026}{Shanghai, China}
\acmBooktitle{Companion of the 2026 ACM International Joint Conference on Pervasive and Ubiquitous Computing (UbiComp Companion '26), October 11--15, 2026, Shanghai, China}
\acmDOI{10.1145/3798063.3839202}
\acmISBN{979-8-4007-2533-3/2026/10}

\begin{document}

\title{HandContextLLM: Task-Aware Evidence Selection for Continuous Hand-Task Understanding and Forecasting}

\author{Chiyue Wang}
\email{202420163875@mail.scut.edu.cn}
\affiliation{%
  \institution{South China University of Technology}
  \city{Guangzhou}
  \country{China}
}

\author{Dong She}
\email{ftdshe@mail.scut.edu.cn}
\affiliation{%
  \institution{South China University of Technology}
  \city{Guangzhou}
  \country{China}
}

\author{Yang Gao}
\authornote{Corresponding author.}
\email{gaoyang2025@scut.edu.cn}
\affiliation{%
  \institution{South China University of Technology}
  \city{Guangzhou}
  \country{China}
}

\author{Zhanpeng Jin}
\email{zjin@scut.edu.cn}
\affiliation{%
  \institution{South China University of Technology}
  \city{Guangzhou}
  \country{China}
}

\begin{abstract}
Interactive wearable agents operating in continuous hand-object tasks must infer not only the current motion, but also its position in the task flow and the likely next action. In this setting, the relevant evidence is query-dependent: current-action understanding, task-stage localization, and next-action forecasting may require different sensor and contextual inputs. We present HandContextLLM, a sensor-language model for continuous hand-task understanding and forecasting. The model combines IMU/sEMG sensor encoders, predicted rolling memory built from its previous outputs, and a frozen Task-Aware Context Selection policy that selects task-relevant evidence according to the query role. We evaluate HandContextLLM on HandContext, a benchmark constructed from self-collected multimodal wearable data and covering current-action understanding, task-stage localization, and conditional next-action forecasting. The results indicate that predicted memory helps recover task progress, sEMG provides useful cues for conditional forecasting, and task-aware evidence selection improves aggregate headline performance relative to fixed input configurations. These findings support sensor-language designs that treat evidence selection as part of task-level context reasoning rather than as a fixed preprocessing choice.

\end{abstract}

\keywords{wearable agents, sensor-language models, hand-task understanding, surface electromyography, inertial sensing, evidence selection, activity forecasting}

\begin{teaserfigure}
  \centering
  \includegraphics[width=\textwidth]{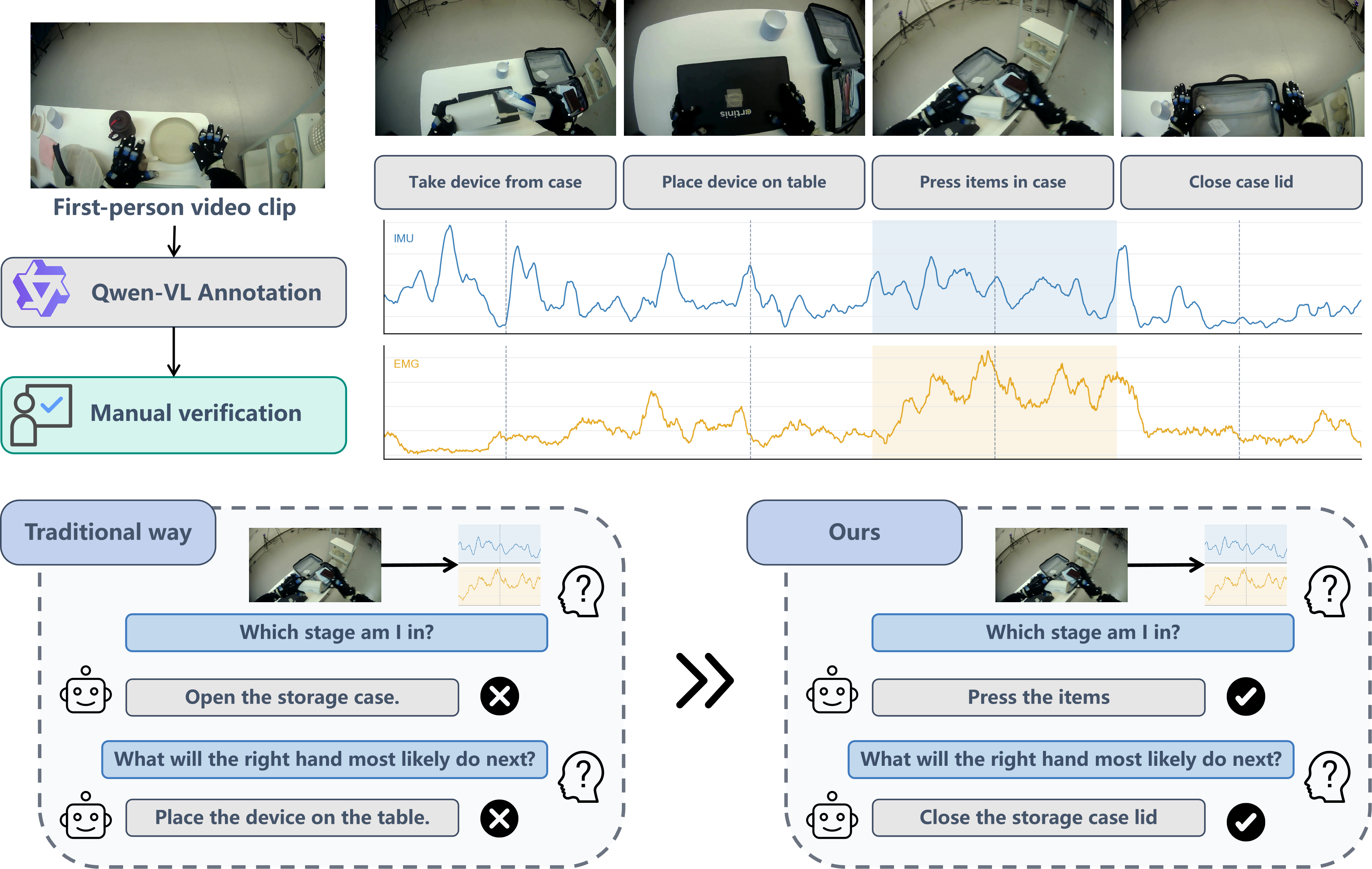}
  \caption{HandContext setting. Video-derived annotations align a continuous hand task with synchronized IMU and sEMG. In this held-out example, fixed context confuses task progress and the next action, whereas HandContextLLM uses task-relevant context to answer both queries correctly. Aggregate results are reported in the evaluation.}
  \Description{A four-step first-person task sequence is aligned with IMU and sEMG signals. A vision-language model and manual verification provide annotations. A fixed-context example gives incorrect stage and next-action answers, while HandContextLLM gives correct answers.}
  \label{fig:overview}
\end{teaserfigure}

\maketitle

\section{Introduction}
\label{sec:introduction}

In continuous hand-object tasks, a wearable agent must do more than assign a label to an isolated motion window. It must answer three related questions: what is happening now, where the action lies in the task, and what is likely to happen next. Window-based activity recognition has developed effective methods for local action classification~\cite{bulling2014tutorial,ordonez2016deepconvlstm}, while systems such as UbiPhysio and ProMemAssist illustrate the value of language feedback and timely proactive assistance~\cite{wang2024ubiphysio,pu2025promemassist}. However, useful assistance also requires connecting local motion to task progress and possible futures.

Recent sensor--language models support richer forms of interaction. Sensor2Text, SensorLLM, and LLaSA align wearable signals with descriptions or question answering~\cite{chen2024sensor2text,li2025sensorllm,imran2024llasa}; SensorQA extends sensor-grounded questions to long daily records~\cite{reichman2025sensorqa}; and HMotionGPT connects continuous hand IMU signals with language~\cite{gao2026hmotiongpt}. Forecasting work further models future wearable activities~\cite{li2025p2lhap}, while neuromotor interfaces and ReactEMG demonstrate the complementary value of surface electromyography (sEMG) for hand interaction and low-latency intent detection~\cite{kaifosh2025neuromotor,wang2025reactemg}. These works, however, do not establish whether queries about the present, task progress, and the future should receive the same evidence.

This paper is based on the observation that the utility of context depends on a query's temporal role. A local window can identify the current action but remain ambiguous about its stage; predicted history can resolve that ambiguity, yet irrelevant or erroneous history can distract forecasting. Likewise, under the same current-action description and scene prior, IMU and sEMG can contribute differently to conditional next-action prediction. Figure~\ref{fig:overview} illustrates a representative setting.

We investigate three research questions:
\begin{enumerate}
  \item[\textbf{RQ1.}] How do IMU and sEMG contribute across current-action understanding, task-progress localization, and conditional next-action forecasting?
  \item[\textbf{RQ2.}] Can past model predictions form an effective rolling memory without test-time ground-truth history?
  \item[\textbf{RQ3.}] Can task-aware evidence selection outperform a fixed input strategy across these tasks?
\end{enumerate}

We present \textbf{HandContextLLM} and the \textbf{HandContext benchmark}, which contains five tasks spanning these three capabilities. Forecasting is explicitly conditioned on the annotated current-action description and scene prior. The model encodes IMU and sEMG windows as sensor tokens and maintains a rolling memory using only its own past action predictions. Motivated by the task-dependency insight above, our \textbf{Task-Aware Context Selection (TACS)} is a deterministic task-level policy fixed before test-time inference: it supplies IMU, sEMG, and predicted memory to state-oriented tasks, while selecting sEMG as the sensor modality for forecasting-oriented tasks.

The evaluation supports this query-dependent design. With textual context held constant, sEMG reaches 95.9\% accuracy on next-action selection and 61.2 ROUGE-L on generation, versus 86.1\% and 32.9 for IMU. Conversely, sEMG alone is weak for progress localization, where adding predicted memory to IMU+sEMG raises accuracy from 27.8\% to 90.4\%. Across one headline metric per task, TACS scores 75.6, compared with 71.9 for fixed predicted memory and 58.5 for fixed sEMG. These comparisons suggest that the gain comes from matching evidence to temporal role, rather than from always adding every available input.

This work makes three contributions:
\begin{enumerate}
  \item We introduce a five-task benchmark that jointly evaluates current action, task progress, and conditional next action, revealing query-dependent roles for sensor and historical evidence.
  \item We introduce predicted rolling memory and evaluate it against no-memory, incorrect-memory, and shuffled-memory controls.
  \item We introduce TACS, a task-level evidence-selection policy that improves balance across task types relative to fixed evidence configurations.
\end{enumerate}

\section{Related Work}
\label{sec:related-work}

Wearable hand sensing commonly relies on inertial measurement units (IMUs) and surface electromyography (sEMG). IMUs capture motion and posture, whereas sEMG reflects muscle activation, providing distinct but complementary evidence~\cite{bulling2014tutorial,zhang2026fusionreview}. Generic wrist-based neuromotor interfaces, ReactEMG, and emg2pose demonstrate the utility of sEMG for continuous control, low-latency action detection, and hand-pose estimation~\cite{kaifosh2025neuromotor,wang2025reactemg,salter2024emg2pose}. ActionSense and MyoKi further synchronize inertial, muscle, video, or kinematic signals during complex and everyday tasks~\cite{delpreto2022actionsense,andreas2025myoki}. These studies establish modality complementarity. We do not aim to develop a new sEMG decoder or generic fusion network; instead, we examine whether each modality has different utility for queries with different temporal roles.

Building on this sensing foundation, wearable activity research has moved from predefined labels toward sensor--language alignment. IMU2CLIP aligns IMU, text, and video; Sensor2Text supports activity captioning and dialogue; and SensorLLM and SensorLM develop language-aligned motion representations at increasing scale~\cite{moon2023imu2clip,chen2024sensor2text,li2025sensorllm,zhang2025sensorlm}. LLaSA and SensorQA extend this direction to open-ended IMU reasoning and questions over long-term daily records~\cite{imran2024llasa,reichman2025sensorqa}. For hand and wrist settings, HMotionGPT aligns continuous bimanual IMU with language for captioning, intent inference, and contextual feedback, while WACU fuses wrist video, audio, and IMU for interaction descriptions~\cite{gao2026hmotiongpt,patsch2025wacu}. Recent EMG work also adopts linguistic representations: AEMG organizes neuromuscular patterns as a self-supervised signal language, and LLM-sEMG maps continuous sEMG into a language-like representation for LLM-based activity recognition~\cite{huang2026aemg,wang2026llmsemg}. These works establish semantic access to hand-sensor data. The question addressed here is whether the sensor and historical evidence supplied to a language model should vary with a query's temporal role.

Once sensor signals become language-model context, continuous tasks also require state tracking and forecasting. Prior work predicts future activities from forecasted IMU signals or jointly performs wearable activity recognition, segmentation, and forecasting~\cite{jaramillo2023activityprediction,li2025p2lhap}. DailyLLM and Health-LLM incorporate multimodal, user, and temporal context into activity logging or health prediction, while ProMemAssist models working memory to time proactive assistance~\cite{tian2025dailyllm,kim2024healthllm,pu2025promemassist}. ZARA further grounds motion time-series reasoning in retrieved discriminative evidence~\cite{li2026zara}. However, these contexts come from structured sensor features, user information, a cognitive working-memory model, or a reference knowledge base, rather than a model's own action predictions during an unfolding task. HandContextLLM uses such predictions as rolling memory and treats history as task-dependent evidence rather than an input that is always beneficial.

Taken together, prior work has advanced multimodal hand sensing, sensor--language interaction, and activity forecasting largely as separate capabilities. HandContextLLM studies their intersection: how sensor modality and self-generated history should be allocated across queries about the present, task progress, and the conditional future.

\section{HandContext Dataset and Benchmark}
\label{sec:benchmark}

\subsection{Data Source and Annotation}
\label{subsec:data-source}

We construct the HandContext benchmark from a self-collected multimodal wearable dataset for continuous hand-object manipulation. Similar to multimodal wearable activity datasets and sensor-language benchmarks~\cite{delpreto2022actionsense,chen2024sensor2text}, HandContext links synchronized sensing streams with interpretable semantic annotations. The dataset contains first-person video, right-hand IMU, and right-hand sEMG from 25 participants across eight long-horizon manipulation scenarios, yielding 194 participant-scene recordings. IMU and sEMG are synchronized at 100~Hz and segmented into sensor windows according to action boundaries. First-person video is used only for offline semantic annotation and as a video reference baseline, not as an inference input to HandContextLLM. Participants provided consent for research use of the collected wearable and video recordings, and the figures in this paper avoid identifiable faces or personal information.

\begin{table}[t]
  \centering
  \caption{Source data and participant-independent split.}
  \label{tab:dataset-summary}
  \small
  \begin{tabular}{lrl}
    \toprule
    Item & Value & Note \\
    \midrule
    Participants & 25 & Split by participant \\
    Scenarios & 8 & Hand-object tasks \\
    Recordings & 194 & 118/16/60 train/val/test \\
    Action annotations & 2,374 & 1,465/190/719 train/val/test \\
    Sensor rate & 100~Hz & Aligned IMU and sEMG \\
    Inference modalities & IMU, sEMG & Video for annotation/baseline \\
    \bottomrule
  \end{tabular}
\end{table}

Action-level text annotations are produced through video-assisted annotation followed by manual verification. Given the first-person video and a predefined prompt, Qwen-VL first produces draft action boundaries, scene priors, and concise action descriptions~\cite{bai2023qwenvl}. Human annotators then correct timestamps, verbs, and object descriptions, and remove ambiguous, heavily occluded, or duplicated segments. Descriptions use a stable operational granularity, focusing on the observable right-hand action and its object rather than unobservable muscle state, force, or subjective intent. The verified text is aligned with IMU/sEMG windows, and the ordered action sequence is used to derive current-action, task-stage, and next-action relations. The source data contains 2,374 action annotations; benchmark instances are derived from these annotations and therefore should not be summed as independent source segments across tasks.

\subsection{Benchmark Tasks}
\label{subsec:benchmark-tasks}

HandContext organizes continuous hand understanding into three capabilities: current-action understanding, task-stage localization, and conditional next-action forecasting. Selection tasks evaluate controlled semantic discrimination, while generation tasks evaluate open-ended action description. Table~\ref{tab:benchmark-tasks} summarizes the five core tasks.

\begin{table}[t]
  \centering
  \caption{HandContext benchmark tasks.}
  \label{tab:benchmark-tasks}
  \scriptsize
  \setlength{\tabcolsep}{2pt}
  \renewcommand{\arraystretch}{1.08}
  \begin{tabular}{@{}p{.20\columnwidth}p{.23\columnwidth}p{.16\columnwidth}p{.15\columnwidth}p{.10\columnwidth}@{}}
    \toprule
    Task & Input & Output & Split & Metric \\
    \midrule
    Current Select & Sensor, scene, opts & Current option & 895/132/462 & Acc. \\
    Current Generate & Sensor, scene & Current text & 895/132/462 & R-L \\
    Stage Locate & Sensor, ordered acts & Stage index & 1,329/201/654 & Acc. \\
    Next Select & Sensor, action, scene, opts & Next option & 670/102/343 & Acc. \\
    Next Generate & Sensor, action, scene & Next text & 670/102/343 & R-L \\
    \bottomrule
  \end{tabular}
\end{table}

Current-Action Description Selection asks the model to choose the action description that best matches the current sensor window from four candidates; Current-Action Description Generation asks it to produce a concise description directly. These tasks test whether local IMU/sEMG evidence can be mapped to action semantics.

Task-Stage Localization provides a current sensor window and three action descriptions from the same task in their true temporal order. The model predicts which stage corresponds to the current window. This is a relative progress-localization task rather than timestamp regression or future prediction; answer positions are kept approximately balanced to reduce position-prior shortcuts.

Next-Action Description Selection and Next-Action Description Generation use instances with valid successor actions. In the formal protocol, the model receives the annotated current-action description and scene prior, so these tasks measure next-action prediction conditioned on current semantics and scene context. In later experiments, ``IMU only'' and ``sEMG only'' refer only to the sensor modality; the textual conditions remain fixed across modality settings.

\subsection{Splits and Evaluation Protocol}
\label{subsec:evaluation-protocol}

We use a participant-independent split: 15/2/8 participants for training, validation, and testing, corresponding to 118/16/60 participant-scene recordings and 1,465/190/719 source action annotations. All benchmark instances are constructed under this split to evaluate generalization to unseen participants and to avoid placing the same person's motion or muscle patterns in both training and testing.

For selection tasks, we report accuracy and macro-F1. For generation tasks, we report ROUGE-L, BLEU-4, and BERTScore-F1~\cite{lin2004rouge,papineni2002bleu,zhang2020bertscore}. The evaluation section highlights accuracy and ROUGE-L as compact headline metrics where needed, while this section defines the full protocol. Overall, HandContext decomposes continuous hand behavior into present semantics, task progress, and conditional future, three linked capabilities that may require different evidence. We next describe HandContextLLM, which selects sensor and historical context according to these task demands.

\section{Method}
\label{sec:method}

\begin{figure*}[t]
  \centering
  \includegraphics[width=\textwidth]{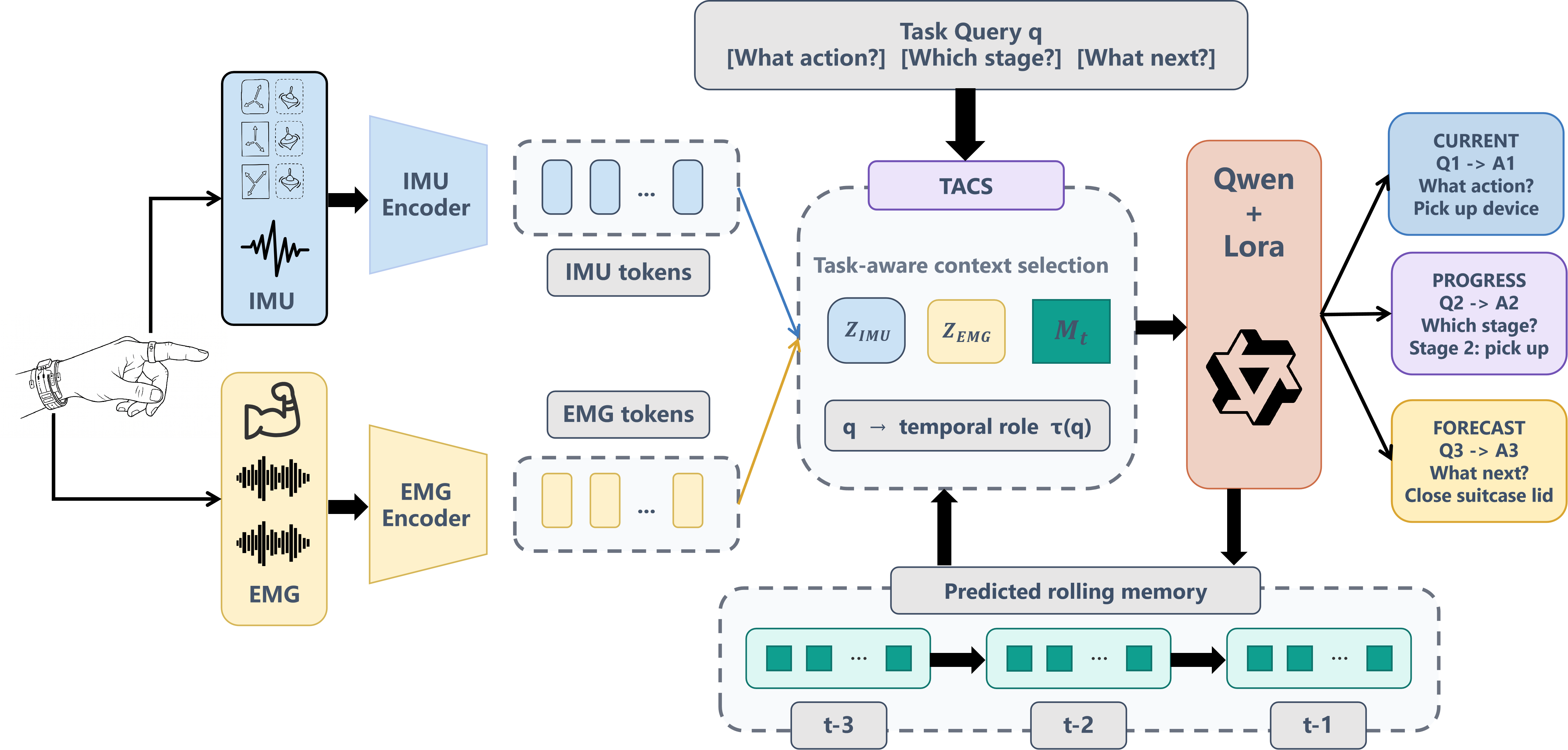}
  \caption{HandContextLLM architecture. IMU and sEMG windows are encoded into sensor tokens. A globally fixed TACS policy, frozen before test-time inference, operationalizes task-aware evidence selection by choosing the context required by each query type: state-oriented queries use IMU, sEMG, and predicted rolling memory, whereas forecasting-oriented queries use sEMG with the same current-action and scene textual conditions. The selected context is decoded by a LoRA-adapted Qwen2.5-7B backbone.}
  \label{fig:architecture}
\end{figure*}

HandContextLLM follows a simple design principle: evidence should match the temporal role of the query. Given a query \(q\), a synchronized sensor window \(\mathbf{x}_t=\{\mathbf{x}_t^{\mathrm{IMU}},\mathbf{x}_t^{\mathrm{sEMG}}\}\), and benchmark-defined textual conditions, the model selects a task-relevant context \(C(q)\) and predicts an answer \(\hat{y}_t\). Figure~\ref{fig:architecture} shows the complete pipeline.

\subsection{Sensor-Language Backbone}
\label{subsec:sensor-language-backbone}

Each modality is first converted into compact sensor tokens:
\begin{equation}
\mathbf{z}_t^{m}=E_m(\mathbf{x}_t^{m}), \quad
m\in\{\mathrm{IMU},\mathrm{sEMG}\}.
\end{equation}
The encoder \(E_m\) uses one-dimensional temporal patching and a lightweight Transformer to summarize within-window dynamics~\cite{vaswani2017attention,nie2023patchtst}. A multi-sensor adapter then fuses the sensor tokens selected by TACS and projects them into the embedding space of Qwen2.5-7B:
\begin{equation}
\mathbf{h}_t=A\left(\{\mathbf{z}_t^m\mid m\in S(q)\}\right).
\end{equation}
This follows the general sensor--language alignment strategy of mapping continuous wearable signals into LLM-compatible representations~\cite{li2025sensorllm,zhang2025sensorlm}; the contribution here is context selection rather than a new time-series foundation encoder. The language model is a frozen Qwen2.5-7B backbone adapted with LoRA~\cite{yang2025qwen25,hu2022lora}. Its prediction is
\begin{equation}
\hat{y}_t
=
F_{\theta_{\mathrm{L}},\theta_{\mathrm{A}}}
\left(\mathbf{p}(q),\mathbf{h}_t,\mathbf{c}_t\right),
\end{equation}
where \(\mathbf{p}(q)\) is the query prompt, \(\mathbf{c}_t\) denotes selected textual context, \(\theta_{\mathrm{L}}\) are LoRA parameters, and \(\theta_{\mathrm{A}}\) are adapter parameters.

\subsection{Predicted Rolling Memory}
\label{subsec:predicted-rolling-memory}

A local sensor window alone does not specify task progress. HandContextLLM therefore maintains a short memory of the model's own previous action predictions:
\begin{equation}
M_t=[\hat{\mathbf{a}}_{t-k},\ldots,\hat{\mathbf{a}}_{t-1}],\quad k=3.
\end{equation}
Each \(\hat{\mathbf{a}}_{t-i}\) is a predicted action card, not raw sensor data and not a ground-truth historical label. In practice, the card contains a concise action description, main verb, object, confidence, and short task-state summary. After each current-action prediction, the memory is updated as
\begin{equation}
M_{t+1}
=
\mathrm{Queue}_k(M_t,\hat{\mathbf{a}}_t),
\end{equation}
where \(\mathrm{Queue}_k\) appends the new prediction and keeps only the latest \(k\) records. This gives the model a compact textual estimate of task progress without using oracle history at test time.

\subsection{Task-Aware Context Selection and Training}
\label{subsec:tacs}

TACS operationalizes task-aware evidence selection as a deterministic task-level policy derived from our task-dependency insight and fixed before test-time inference for all test samples. It uses only the query type, not test labels or per-instance outcomes. It is not a learned router and is not selected per test instance. We group current-action understanding and task-stage localization as state-oriented queries \(\mathcal{Q}_{\mathrm{state}}\), and next-action tasks as forecasting-oriented queries \(\mathcal{Q}_{\mathrm{forecast}}\). TACS selects
\begin{equation}
C(q)=
\begin{cases}
\{\mathbf{z}_{t}^{\mathrm{IMU}},\mathbf{z}_{t}^{\mathrm{sEMG}},M_t\},
& q\in\mathcal{Q}_{\mathrm{state}},\\
\{\mathbf{z}_{t}^{\mathrm{sEMG}},d_t,s\},
& q\in\mathcal{Q}_{\mathrm{forecast}},
\end{cases}
\end{equation}
where \(d_t\) is the current-action textual condition and \(s\) is the scene prior. Thus, state-oriented queries receive multimodal local evidence plus predicted progress history, while forecasting queries compare sensor evidence under the same current-action and scene conditions.

Training optimizes only the sensor adapter and LoRA parameters; the Qwen2.5-7B backbone remains frozen:
\begin{equation}
(\theta_{\mathrm{A}}^{*},\theta_{\mathrm{L}}^{*})
=
\arg\min_{\theta_{\mathrm{A}},\theta_{\mathrm{L}}}
\sum_{(q,\mathbf{x},\mathbf{c},y)\in\mathcal{D}}
\mathcal{L}\!\left(
F_{\theta_{\mathrm{L}},\theta_{\mathrm{A}}}
(\mathbf{p}(q),\mathbf{h},\mathbf{c}),y
\right).
\end{equation}
For selection tasks, \(\mathcal{L}\) is the cross-entropy loss over answer choices; for generation tasks, it is the token-level negative log-likelihood of the target action description. At inference time, the frozen TACS policy chooses \(C(q)\), the model predicts \(\hat{y}_t\), and current-action outputs update \(M_t\) for later state-oriented queries. This formulation also defines the fixed-configuration and memory-control variants used in the evaluation.

\section{Evaluation}
\label{sec:evaluation}

\subsection{Experimental Setup}
\label{subsec:eval-setup}

We use the participant-independent test split and metrics defined in Section~\ref{sec:benchmark}. Tables report all task metrics: accuracy and macro-F1 for selection tasks, and ROUGE-L, BLEU-4, and BERTScore-F1 for generation tasks. For compact comparison, we also report a headline average over accuracy for selection tasks and ROUGE-L for generation tasks. Let \(\mathcal{T}_{h}\) be the five headline tasks and \(s_t(m)\) the corresponding headline score. We define
\begin{equation}
\mathrm{CoreAvg}(m)=\frac{1}{|\mathcal{T}_{h}|}
\sum_{t\in\mathcal{T}_{h}} s_t(m).
\end{equation}
Except where noted, \(mean\pm std\) is computed from 100 rounds of 90\% test-set subsampling.

\newcommand{\hcmetricpair}[2]{\begin{tabular}[t]{@{}l@{}}#1\\#2\end{tabular}}

\begin{figure}[t]
  \centering
  \includegraphics[width=.95\columnwidth]{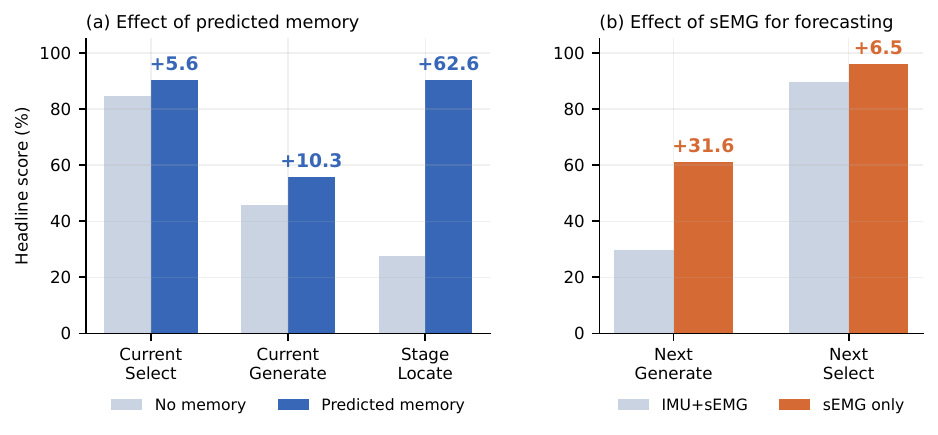}
  \caption{Task-dependent evidence patterns. Predicted memory improves state-oriented tasks, while sEMG better supports conditional forecasting. Selection tasks use accuracy; generation tasks use ROUGE-L.}
  \label{fig:task-dependency}
\end{figure}

\subsection{Overall Performance}
\label{subsec:overall-performance}

Table~\ref{tab:overall-performance} compares HandContextLLM with LLMs that directly read compressed IMU/sEMG text. HandContextLLM is best on all five headline metrics, with 78.7 CoreAvg versus 51.4 for the strongest baseline. This gap indicates that continuous hand-task understanding benefits from learned sensor representations and task-relevant context.

\begin{table}[t]
  \centering
  \caption{Overall performance. Selection task cells report Acc. over F1; generation task cells report ROUGE-L/BLEU-4/BERTScore-F1. CoreAvg averages one headline metric per task: selection accuracy and generation ROUGE-L.}
  \label{tab:overall-performance}
  \scriptsize
  \setlength{\tabcolsep}{2pt}
  \renewcommand{\arraystretch}{1.08}
  \textit{Selection tasks and headline average.}\par\vspace{1pt}
  \begin{tabular}{@{}p{.25\columnwidth}p{.18\columnwidth}p{.18\columnwidth}p{.18\columnwidth}p{.12\columnwidth}@{}}
    \toprule
    Method & Current & Stage & Next & Core \\
    \midrule
    Local Qwen2.5 & \hcmetricpair{54.3$\pm$0.7}{51.7$\pm$0.7} & \hcmetricpair{33.6$\pm$0.6}{25.9$\pm$0.5} & \hcmetricpair{86.9$\pm$0.6}{86.8$\pm$0.6} & 44.5$\pm$0.2 \\
    Local Qwen3 & \hcmetricpair{50.0$\pm$0.7}{49.3$\pm$0.7} & \hcmetricpair{33.3$\pm$0.6}{27.5$\pm$0.5} & \hcmetricpair{75.2$\pm$0.8}{75.1$\pm$0.8} & 47.4$\pm$0.3 \\
    Qwen-Flash & \hcmetricpair{56.0$\pm$0.8}{56.0$\pm$0.8} & \hcmetricpair{34.2$\pm$0.6}{25.7$\pm$0.5} & \hcmetricpair{84.8$\pm$0.6}{84.5$\pm$0.6} & 51.4$\pm$0.2 \\
    Kimi-K2.5 & \hcmetricpair{45.6$\pm$0.8}{49.0$\pm$0.8} & \hcmetricpair{30.4$\pm$0.5}{26.2$\pm$0.6} & \hcmetricpair{85.1$\pm$0.6}{86.1$\pm$0.6} & 48.0$\pm$0.2 \\
    DeepSeek & \hcmetricpair{52.4$\pm$0.8}{51.5$\pm$0.8} & \hcmetricpair{33.7$\pm$0.7}{25.9$\pm$0.5} & \hcmetricpair{85.4$\pm$0.6}{85.2$\pm$0.6} & 50.6$\pm$0.2 \\
    Gemini-Flash-Lite & \hcmetricpair{51.0$\pm$0.9}{50.0$\pm$0.9} & \hcmetricpair{33.3$\pm$0.6}{29.6$\pm$0.6} & \hcmetricpair{84.2$\pm$0.6}{84.0$\pm$0.6} & 45.5$\pm$0.3 \\
    HandContextLLM & \textbf{\hcmetricpair{90.2$\pm$0.5}{90.5$\pm$0.5}} & \textbf{\hcmetricpair{90.4$\pm$0.4}{90.5$\pm$0.4}} & \textbf{\hcmetricpair{95.9$\pm$0.3}{96.0$\pm$0.3}} & \textbf{78.7$\pm$0.2} \\
    \bottomrule
  \end{tabular}
  \par\vspace{3pt}
  \textit{Generation tasks.}\par\vspace{1pt}
  \begin{tabular}{@{}p{.25\columnwidth}p{.33\columnwidth}p{.33\columnwidth}@{}}
    \toprule
    Method & Current & Next \\
    \midrule
    Local Qwen2.5 & 7.1$\pm$0.2 / 0.1$\pm$0.0 / 62.3$\pm$0.1 & 40.7$\pm$0.3 / 9.5$\pm$0.3 / 77.4$\pm$0.1 \\
    Local Qwen3 & 30.1$\pm$0.2 / 0.2$\pm$0.0 / 73.0$\pm$0.1 & 48.3$\pm$0.3 / 18.1$\pm$0.5 / 80.3$\pm$0.1 \\
    Qwen-Flash & 30.9$\pm$0.2 / 0.2$\pm$0.0 / 75.0$\pm$0.1 & 50.9$\pm$0.4 / 20.8$\pm$0.4 / 81.0$\pm$0.1 \\
    Kimi-K2.5 & 29.9$\pm$0.2 / 4.0$\pm$0.2 / 72.8$\pm$0.1 & 48.9$\pm$0.3 / 18.6$\pm$0.4 / 79.7$\pm$0.1 \\
    DeepSeek & 34.1$\pm$0.2 / 7.8$\pm$0.2 / 74.0$\pm$0.1 & 47.6$\pm$0.3 / 18.3$\pm$0.4 / 79.3$\pm$0.1 \\
    Gemini-Flash-Lite & 25.1$\pm$0.2 / 0.9$\pm$0.1 / 69.7$\pm$0.1 & 34.0$\pm$0.3 / 8.2$\pm$0.4 / 74.0$\pm$0.1 \\
    HandContextLLM & \textbf{55.8$\pm$0.5 / 32.7$\pm$0.6 / 79.3$\pm$0.3} & \textbf{61.4$\pm$0.5 / 40.7$\pm$0.6 / 83.8$\pm$0.2} \\
    \bottomrule
  \end{tabular}
\end{table}

\subsection{Modality Ablation}
\label{subsec:modality-ablation}

Table~\ref{tab:modality-ablation} shows clear task dependency. sEMG-only is strongest for conditional forecasting, reaching 61.2 ROUGE-L on Next-Generate and 95.9 accuracy on Next-Select. Yet it reaches only 11.6 accuracy on Stage-Locate, indicating that progress localization cannot be solved from the current sEMG window alone. Non-wearable or hard-to-deploy reference modalities are included only as diagnostic references.

\begin{table}[t]
  \centering
  \caption{Modality ablation and reference modalities. Selection task cells report Acc. over F1; generation task cells report ROUGE-L/BLEU-4/BERTScore-F1. CoreAvg averages one headline metric per task: selection accuracy and generation ROUGE-L.}
  \label{tab:modality-ablation}
  \scriptsize
  \setlength{\tabcolsep}{2pt}
  \renewcommand{\arraystretch}{1.08}
  \textit{Selection tasks and headline average.}\par\vspace{1pt}
  \begin{tabular}{@{}p{.25\columnwidth}p{.18\columnwidth}p{.18\columnwidth}p{.18\columnwidth}p{.12\columnwidth}@{}}
    \toprule
    Method & Current & Stage & Next & Core \\
    \midrule
    IMU only & \hcmetricpair{60.4$\pm$0.7}{72.9$\pm$0.6} & \hcmetricpair{27.3$\pm$0.6}{24.7$\pm$0.5} & \hcmetricpair{86.1$\pm$0.6}{86.9$\pm$0.6} & 50.0$\pm$0.2 \\
    sEMG only & \hcmetricpair{83.0$\pm$0.5}{86.0$\pm$0.5} & \hcmetricpair{11.6$\pm$0.4}{11.5$\pm$0.3} & \textbf{\hcmetricpair{95.9$\pm$0.4}{96.0$\pm$0.4}} & 59.0$\pm$0.2 \\
    EyeTrack only & \hcmetricpair{86.2$\pm$0.5}{87.3$\pm$0.5} & \hcmetricpair{19.9$\pm$0.5}{19.8$\pm$0.5} & \hcmetricpair{34.7$\pm$0.9}{50.8$\pm$1.0} & 44.2$\pm$0.3 \\
    IMU+sEMG & \hcmetricpair{84.7$\pm$0.6}{85.2$\pm$0.6} & \hcmetricpair{28.0$\pm$0.6}{25.5$\pm$0.6} & \hcmetricpair{89.5$\pm$0.6}{90.9$\pm$0.6} & 55.5$\pm$0.2 \\
    Pressure only & \hcmetricpair{80.0$\pm$0.6}{80.2$\pm$0.6} & \hcmetricpair{17.6$\pm$0.5}{21.4$\pm$0.6} & \hcmetricpair{86.5$\pm$0.6}{89.7$\pm$0.5} & 51.0$\pm$0.2 \\
    Mocap only & \hcmetricpair{82.3$\pm$0.6}{82.7$\pm$0.6} & \hcmetricpair{31.1$\pm$0.6}{27.3$\pm$0.5} & \hcmetricpair{88.9$\pm$0.5}{91.6$\pm$0.4} & 55.8$\pm$0.3 \\
    Video only & \hcmetricpair{72.5$\pm$1.3}{70.6$\pm$1.4} & \hcmetricpair{38.7$\pm$0.8}{31.6$\pm$0.6} & \hcmetricpair{81.4$\pm$1.1}{81.5$\pm$1.2} & 53.7$\pm$0.4 \\
    HandContextLLM & \textbf{\hcmetricpair{90.3$\pm$0.5}{90.5$\pm$0.5}} & \textbf{\hcmetricpair{90.4$\pm$0.4}{90.4$\pm$0.3}} & \textbf{\hcmetricpair{95.9$\pm$0.4}{96.0$\pm$0.4}} & \textbf{78.8$\pm$0.2} \\
    \bottomrule
  \end{tabular}
  \par\vspace{3pt}
  \textit{Generation tasks.}\par\vspace{1pt}
  \begin{tabular}{@{}p{.25\columnwidth}p{.33\columnwidth}p{.33\columnwidth}@{}}
    \toprule
    Method & Current & Next \\
    \midrule
    IMU only & 43.1$\pm$0.3 / 11.4$\pm$0.3 / 77.5$\pm$0.1 & 32.9$\pm$0.5 / 20.1$\pm$0.5 / 47.9$\pm$0.7 \\
    sEMG only & 43.3$\pm$0.3 / 12.7$\pm$0.4 / 76.4$\pm$0.2 & 61.2$\pm$0.6 / 40.5$\pm$0.7 / 83.7$\pm$0.3 \\
    EyeTrack only & 36.2$\pm$0.4 / 11.1$\pm$0.3 / 67.0$\pm$0.4 & 44.3$\pm$0.5 / 27.3$\pm$0.6 / 62.5$\pm$0.6 \\
    IMU+sEMG & 45.6$\pm$0.4 / 13.7$\pm$0.4 / 78.7$\pm$0.2 & 29.6$\pm$0.6 / 17.7$\pm$0.6 / 55.4$\pm$0.6 \\
    Pressure only & 40.1$\pm$0.4 / 12.9$\pm$0.4 / 71.6$\pm$0.3 & 30.9$\pm$0.5 / 15.4$\pm$0.4 / 65.4$\pm$0.4 \\
    Mocap only & 40.6$\pm$0.4 / 11.8$\pm$0.3 / 75.8$\pm$0.2 & 36.0$\pm$0.6 / 21.0$\pm$0.5 / 55.2$\pm$0.7 \\
    Video only & 38.0$\pm$0.5 / 11.2$\pm$0.4 / 75.6$\pm$0.2 & 37.9$\pm$0.5 / 7.5$\pm$0.5 / 78.6$\pm$0.2 \\
    HandContextLLM & \textbf{55.8$\pm$0.4 / 32.7$\pm$0.5 / 79.3$\pm$0.3} & \textbf{61.3$\pm$0.5 / 40.6$\pm$0.7 / 83.8$\pm$0.2} \\
    \bottomrule
  \end{tabular}
\end{table}

\subsection{Memory Ablation and TACS}
\label{subsec:memory-tacs}

With TACS, Stage-Locate reaches 90.3 versus 27.7 without memory. Wrong memory lowers CoreAvg to 45.7, while shuffled memory helps Stage-Locate but harms Next-Select, showing that temporally mismatched memory can harm.

\begin{table}[t]
  \centering
  \caption{Memory and context-selection ablation. Selection task cells report Acc. over F1; generation task cells report ROUGE-L/BLEU-4/BERTScore-F1. CoreAvg averages one headline metric per task: selection accuracy and generation ROUGE-L.}
  \label{tab:memory-tacs}
  \scriptsize
  \setlength{\tabcolsep}{2pt}
  \renewcommand{\arraystretch}{1.08}
  \textit{Selection tasks and headline average.}\par\vspace{1pt}
  \begin{tabular}{@{}p{.25\columnwidth}p{.18\columnwidth}p{.18\columnwidth}p{.18\columnwidth}p{.12\columnwidth}@{}}
    \toprule
    Method & Current & Stage & Next & Core \\
    \midrule
    No memory & \hcmetricpair{84.7$\pm$0.5}{85.1$\pm$0.5} & \hcmetricpair{27.7$\pm$0.6}{25.3$\pm$0.6} & \hcmetricpair{89.4$\pm$0.6}{90.9$\pm$0.6} & 55.4$\pm$0.2 \\
    Wrong memory & \hcmetricpair{80.0$\pm$0.6}{83.3$\pm$0.6} & \hcmetricpair{30.7$\pm$0.6}{28.0$\pm$0.6} & \hcmetricpair{80.6$\pm$0.7}{81.8$\pm$0.6} & 45.7$\pm$0.3 \\
    Shuffled memory & \hcmetricpair{89.3$\pm$0.4}{89.8$\pm$0.4} & \hcmetricpair{80.9$\pm$0.5}{86.0$\pm$0.4} & \hcmetricpair{70.3$\pm$0.9}{72.7$\pm$0.9} & 65.7$\pm$0.3 \\
    HandContextLLM & \textbf{\hcmetricpair{90.3$\pm$0.6}{90.5$\pm$0.6}} & \textbf{\hcmetricpair{90.3$\pm$0.4}{90.4$\pm$0.4}} & \textbf{\hcmetricpair{96.0$\pm$0.4}{96.1$\pm$0.4}} & \textbf{78.7$\pm$0.2} \\
    \bottomrule
  \end{tabular}
  \par\vspace{3pt}
  \textit{Generation tasks.}\par\vspace{1pt}
  \begin{tabular}{@{}p{.25\columnwidth}p{.33\columnwidth}p{.33\columnwidth}@{}}
    \toprule
    Method & Current & Next \\
    \midrule
    No memory & 45.6$\pm$0.4 / 13.6$\pm$0.5 / 78.7$\pm$0.2 & 29.6$\pm$0.6 / 17.8$\pm$0.5 / 55.4$\pm$0.6 \\
    Wrong memory & 9.2$\pm$0.3 / 2.2$\pm$0.2 / 18.0$\pm$0.5 & 28.0$\pm$0.6 / 15.4$\pm$0.5 / 55.5$\pm$0.6 \\
    Shuffled memory & 34.2$\pm$0.5 / 18.3$\pm$0.5 / 70.8$\pm$0.3 & 53.7$\pm$0.6 / 34.4$\pm$0.7 / 77.7$\pm$0.4 \\
    HandContextLLM & \textbf{55.9$\pm$0.4 / 32.7$\pm$0.5 / 79.3$\pm$0.3} & \textbf{61.2$\pm$0.5 / 40.5$\pm$0.7 / 83.7$\pm$0.3} \\
    \bottomrule
  \end{tabular}
\end{table}

TACS reaches 75.6 CoreAvg versus 71.9 for fixed IMU+sEMG+predicted memory.

\section{Discussion}
\label{sec:discussion}

On HandContext, current-action understanding, task-stage localization, and next-action forecasting require different evidence: predicted memory helps recover task progress, whereas sEMG aids conditional forecasting given current-action text and scene context.

Predicted rolling memory uses model predictions rather than ground-truth history; wrong-memory and shuffled-memory controls show that inaccurate history can mislead. TACS is a frozen task-level policy: state-oriented queries use IMU, sEMG, and predicted memory; forecasting queries use sEMG with benchmark text. It is neither learned per instance nor adaptive to evidence shifts or signal-quality changes.

This offline benchmark does not measure system latency, resource use, user interaction, or recovery from early memory errors. Distillation and quantization remain practical directions.

\section{Conclusion}
\label{sec:conclusion}

We presented HandContextLLM, which combines IMU/sEMG encoding, predicted rolling memory, and a frozen task-aware context policy. On HandContext, predicted memory improves state-oriented tasks, sEMG aids conditional next-action forecasting, and the policy improves headline performance. These benchmark results suggest that task-progress context can complement a single sensor window.

\clearpage
\bibliographystyle{ACM-Reference-Format}
\bibliography{references}


\begin{thebibliography}{33}


\ifx \showCODEN    \undefined \def \showCODEN     #1{\unskip}     \fi
\ifx \showISBNx    \undefined \def \showISBNx     #1{\unskip}     \fi
\ifx \showISBNxiii \undefined \def \showISBNxiii  #1{\unskip}     \fi
\ifx \showISSN     \undefined \def \showISSN      #1{\unskip}     \fi
\ifx \showLCCN     \undefined \def \showLCCN      #1{\unskip}     \fi
\ifx \shownote     \undefined \def \shownote      #1{#1}          \fi
\ifx \showarticletitle \undefined \def \showarticletitle #1{#1}   \fi
\ifx \showURL      \undefined \def \showURL       {\relax}        \fi
\providecommand\bibfield[2]{#2}
\providecommand\bibinfo[2]{#2}
\providecommand\natexlab[1]{#1}
\providecommand\showeprint[2][]{arXiv:#2}

\bibitem[Andreas et~al\mbox{.}(2025)]%
        {andreas2025myoki}
\bibfield{author}{\bibinfo{person}{Daniel Andreas}, \bibinfo{person}{Dominik Werner}, \bibinfo{person}{Zhongshi Hou}, \bibinfo{person}{Anany Dwivedi}, \bibinfo{person}{Claudio Castellini}, {and} \bibinfo{person}{Philipp Beckerle}.} \bibinfo{year}{2025}\natexlab{}.
\newblock \showarticletitle{A Database of Multimodal Myography and Hand Kinematics During Realistic Daily Life Activities}.
\newblock \bibinfo{journal}{\emph{Scientific Data}}  \bibinfo{volume}{12}, Article \bibinfo{articleno}{1507} (\bibinfo{year}{2025}), \bibinfo{numpages}{14}~pages.
\newblock
\href{https://doi.org/10.1038/s41597-025-05852-6}{doi:\nolinkurl{10.1038/s41597-025-05852-6}}


\bibitem[Bai et~al\mbox{.}(2023)]%
        {bai2023qwenvl}
\bibfield{author}{\bibinfo{person}{Jinze Bai}, \bibinfo{person}{Shuai Bai}, \bibinfo{person}{Shusheng Yang}, \bibinfo{person}{Shijie Wang}, \bibinfo{person}{Sinan Tan}, \bibinfo{person}{Peng Wang}, \bibinfo{person}{Junyang Lin}, \bibinfo{person}{Chang Zhou}, {and} \bibinfo{person}{Jingren Zhou}.} \bibinfo{year}{2023}\natexlab{}.
\newblock \showarticletitle{{Qwen-VL}: A Versatile Vision-Language Model for Understanding, Localization, Text Reading, and Beyond}.
\newblock \bibinfo{journal}{\emph{arXiv preprint arXiv:2308.12966}} (\bibinfo{year}{2023}).
\newblock
\urldef\tempurl%
\url{https://arxiv.org/abs/2308.12966}
\showURL{%
\tempurl}


\bibitem[Bulling et~al\mbox{.}(2014)]%
        {bulling2014tutorial}
\bibfield{author}{\bibinfo{person}{Andreas Bulling}, \bibinfo{person}{Ulf Blanke}, {and} \bibinfo{person}{Bernt Schiele}.} \bibinfo{year}{2014}\natexlab{}.
\newblock \showarticletitle{A Tutorial on Human Activity Recognition Using Body-Worn Inertial Sensors}.
\newblock \bibinfo{journal}{\emph{Comput. Surveys}} \bibinfo{volume}{46}, \bibinfo{number}{3}, Article \bibinfo{articleno}{33} (\bibinfo{year}{2014}), \bibinfo{numpages}{33}~pages.
\newblock
\href{https://doi.org/10.1145/2499621}{doi:\nolinkurl{10.1145/2499621}}


\bibitem[Chen et~al\mbox{.}(2024)]%
        {chen2024sensor2text}
\bibfield{author}{\bibinfo{person}{Wenqiang Chen}, \bibinfo{person}{Jiaxuan Cheng}, \bibinfo{person}{Leyao Wang}, \bibinfo{person}{Wei Zhao}, {and} \bibinfo{person}{Wojciech Matusik}.} \bibinfo{year}{2024}\natexlab{}.
\newblock \showarticletitle{Sensor2Text: Enabling Natural Language Interactions for Daily Activity Tracking Using Wearable Sensors}.
\newblock \bibinfo{journal}{\emph{Proceedings of the ACM on Interactive, Mobile, Wearable and Ubiquitous Technologies}} \bibinfo{volume}{8}, \bibinfo{number}{4}, Article \bibinfo{articleno}{192} (\bibinfo{year}{2024}), \bibinfo{numpages}{26}~pages.
\newblock
\href{https://doi.org/10.1145/3699747}{doi:\nolinkurl{10.1145/3699747}}


\bibitem[DelPreto et~al\mbox{.}(2022)]%
        {delpreto2022actionsense}
\bibfield{author}{\bibinfo{person}{Joseph DelPreto}, \bibinfo{person}{Chao Liu}, \bibinfo{person}{Yiyue Luo}, \bibinfo{person}{Michael Foshey}, \bibinfo{person}{Yunzhu Li}, \bibinfo{person}{Antonio Torralba}, \bibinfo{person}{Wojciech Matusik}, {and} \bibinfo{person}{Daniela Rus}.} \bibinfo{year}{2022}\natexlab{}.
\newblock \showarticletitle{{ActionSense}: A Multimodal Dataset and Recording Framework for Human Activities Using Wearable Sensors in a Kitchen Environment}. In \bibinfo{booktitle}{\emph{Advances in Neural Information Processing Systems}}, Vol.~\bibinfo{volume}{35}. \bibinfo{pages}{13800--13813}.
\newblock
\urldef\tempurl%
\url{https://proceedings.neurips.cc/paper_files/paper/2022/hash/5985e81d65605827ac35401999aea22a-Abstract-Datasets_and_Benchmarks.html}
\showURL{%
\tempurl}


\bibitem[Gao et~al\mbox{.}(2026)]%
        {gao2026hmotiongpt}
\bibfield{author}{\bibinfo{person}{Yang Gao}, \bibinfo{person}{Dong She}, \bibinfo{person}{Wolin Liang}, \bibinfo{person}{Chiyue Wang}, \bibinfo{person}{Yingjing Xiao}, \bibinfo{person}{Xianrong Yao}, \bibinfo{person}{Cong Liu}, \bibinfo{person}{Zhichao Huang}, {and} \bibinfo{person}{Zhanpeng Jin}.} \bibinfo{year}{2026}\natexlab{}.
\newblock \showarticletitle{{HMotionGPT}: Aligning Hand Motions and Natural Language for Activity Understanding with Smart Rings}.
\newblock \bibinfo{journal}{\emph{Proceedings of the ACM on Interactive, Mobile, Wearable and Ubiquitous Technologies}} \bibinfo{volume}{10}, \bibinfo{number}{2}, Article \bibinfo{articleno}{37} (\bibinfo{year}{2026}), \bibinfo{numpages}{33}~pages.
\newblock
\href{https://doi.org/10.1145/3810222}{doi:\nolinkurl{10.1145/3810222}}


\bibitem[Hu et~al\mbox{.}(2022)]%
        {hu2022lora}
\bibfield{author}{\bibinfo{person}{Edward~J. Hu}, \bibinfo{person}{Yelong Shen}, \bibinfo{person}{Phillip Wallis}, \bibinfo{person}{Zeyuan Allen-Zhu}, \bibinfo{person}{Yuanzhi Li}, \bibinfo{person}{Shean Wang}, \bibinfo{person}{Lu Wang}, {and} \bibinfo{person}{Weizhu Chen}.} \bibinfo{year}{2022}\natexlab{}.
\newblock \showarticletitle{{LoRA}: Low-Rank Adaptation of Large Language Models}. In \bibinfo{booktitle}{\emph{International Conference on Learning Representations}}.
\newblock
\urldef\tempurl%
\url{https://openreview.net/forum?id=nZeVKeeFYf9}
\showURL{%
\tempurl}


\bibitem[Huang et~al\mbox{.}(2026)]%
        {huang2026aemg}
\bibfield{author}{\bibinfo{person}{Zhenghao Huang}, \bibinfo{person}{Huilin Yao}, \bibinfo{person}{Kaikai Wang}, {and} \bibinfo{person}{Lin Shu}.} \bibinfo{year}{2026}\natexlab{}.
\newblock \showarticletitle{Reading Your Actions: Learning Generalizable Action Representations via Pre-training {AEMG}}. In \bibinfo{booktitle}{\emph{Proceedings of the IEEE/CVF Conference on Computer Vision and Pattern Recognition}}. \bibinfo{publisher}{IEEE}, \bibinfo{address}{Piscataway, NJ, USA}, \bibinfo{pages}{20108--20117}.
\newblock
\urldef\tempurl%
\url{https://openaccess.thecvf.com/content/CVPR2026/html/Huang_Reading_Your_Actions_Learning_Generalizable_Action_Representations_via_Pre-training_AEMG_CVPR_2026_paper.html}
\showURL{%
\tempurl}


\bibitem[Imran et~al\mbox{.}(2024)]%
        {imran2024llasa}
\bibfield{author}{\bibinfo{person}{Sheikh~Asif Imran}, \bibinfo{person}{Mohammad Nur~Hossain Khan}, \bibinfo{person}{Subrata Biswas}, {and} \bibinfo{person}{Bashima Islam}.} \bibinfo{year}{2024}\natexlab{}.
\newblock \showarticletitle{{LLaSA}: A Sensor-Aware {LLM} for Natural Language Reasoning of Human Activity from {IMU} Data}.
\newblock \bibinfo{journal}{\emph{arXiv preprint arXiv:2406.14498}} (\bibinfo{year}{2024}).
\newblock
\urldef\tempurl%
\url{https://arxiv.org/abs/2406.14498}
\showURL{%
\tempurl}


\bibitem[Jaramillo et~al\mbox{.}(2023)]%
        {jaramillo2023activityprediction}
\bibfield{author}{\bibinfo{person}{Ismael~Espinoza Jaramillo}, \bibinfo{person}{Channabasava Chola}, \bibinfo{person}{Jin-Gyun Jeong}, \bibinfo{person}{Ji-Heon Oh}, \bibinfo{person}{Hwanseok Jung}, \bibinfo{person}{Jin-Hyuk Lee}, \bibinfo{person}{Won~Hee Lee}, {and} \bibinfo{person}{Tae-Seong Kim}.} \bibinfo{year}{2023}\natexlab{}.
\newblock \showarticletitle{Human Activity Prediction Based on Forecasted {IMU} Activity Signals by Sequence-to-Sequence Deep Neural Networks}.
\newblock \bibinfo{journal}{\emph{Sensors}} \bibinfo{volume}{23}, \bibinfo{number}{14} (\bibinfo{year}{2023}), \bibinfo{pages}{6491}.
\newblock
\href{https://doi.org/10.3390/s23146491}{doi:\nolinkurl{10.3390/s23146491}}


\bibitem[Kaifosh et~al\mbox{.}(2025)]%
        {kaifosh2025neuromotor}
\bibfield{author}{\bibinfo{person}{Patrick Kaifosh}, \bibinfo{person}{Thomas~R. Reardon}, {and} \bibinfo{person}{{CTRL-labs at Reality Labs}}.} \bibinfo{year}{2025}\natexlab{}.
\newblock \showarticletitle{A Generic Non-Invasive Neuromotor Interface for Human--Computer Interaction}.
\newblock \bibinfo{journal}{\emph{Nature}}  \bibinfo{volume}{645} (\bibinfo{year}{2025}), \bibinfo{pages}{702--711}.
\newblock
\href{https://doi.org/10.1038/s41586-025-09255-w}{doi:\nolinkurl{10.1038/s41586-025-09255-w}}


\bibitem[Kim et~al\mbox{.}(2024)]%
        {kim2024healthllm}
\bibfield{author}{\bibinfo{person}{Yubin Kim}, \bibinfo{person}{Xuhai Xu}, \bibinfo{person}{Daniel McDuff}, \bibinfo{person}{Cynthia Breazeal}, {and} \bibinfo{person}{Hae~Won Park}.} \bibinfo{year}{2024}\natexlab{}.
\newblock \showarticletitle{Health-{LLM}: Large Language Models for Health Prediction via Wearable Sensor Data}. In \bibinfo{booktitle}{\emph{Proceedings of the Fifth Conference on Health, Inference, and Learning}} \emph{(\bibinfo{series}{Proceedings of Machine Learning Research}, Vol.~\bibinfo{volume}{248})}. \bibinfo{publisher}{PMLR}, \bibinfo{address}{United States}, \bibinfo{pages}{522--539}.
\newblock
\urldef\tempurl%
\url{https://proceedings.mlr.press/v248/kim24b.html}
\showURL{%
\tempurl}


\bibitem[Li et~al\mbox{.}(2025b)]%
        {li2025p2lhap}
\bibfield{author}{\bibinfo{person}{Shuangjian Li}, \bibinfo{person}{Tao Zhu}, \bibinfo{person}{Mingxing Nie}, \bibinfo{person}{Huansheng Ning}, \bibinfo{person}{Zhenyu Liu}, {and} \bibinfo{person}{Liming Chen}.} \bibinfo{year}{2025}\natexlab{b}.
\newblock \showarticletitle{{P2LHAP}: Wearable-Sensor-Based Human Activity Recognition, Segmentation, and Forecast Through Patch-to-Label {Seq2Seq} Transformer}.
\newblock \bibinfo{journal}{\emph{IEEE Internet of Things Journal}} \bibinfo{volume}{12}, \bibinfo{number}{6} (\bibinfo{year}{2025}), \bibinfo{pages}{6818--6830}.
\newblock
\href{https://doi.org/10.1109/JIOT.2024.3493380}{doi:\nolinkurl{10.1109/JIOT.2024.3493380}}


\bibitem[Li et~al\mbox{.}(2026)]%
        {li2026zara}
\bibfield{author}{\bibinfo{person}{Zechen Li}, \bibinfo{person}{Baiyu Chen}, \bibinfo{person}{Hao Xue}, {and} \bibinfo{person}{Flora~D. Salim}.} \bibinfo{year}{2026}\natexlab{}.
\newblock \showarticletitle{{ZARA}: Training-Free Motion Time-Series Reasoning via Evidence-Grounded {LLM} Agents}. In \bibinfo{booktitle}{\emph{Proceedings of the 64th Annual Meeting of the Association for Computational Linguistics (Volume 1: Long Papers)}}. \bibinfo{publisher}{Association for Computational Linguistics}, \bibinfo{address}{San Diego, California, United States}, \bibinfo{pages}{14986--15008}.
\newblock
\href{https://doi.org/10.18653/v1/2026.acl-long.684}{doi:\nolinkurl{10.18653/v1/2026.acl-long.684}}


\bibitem[Li et~al\mbox{.}(2025a)]%
        {li2025sensorllm}
\bibfield{author}{\bibinfo{person}{Zechen Li}, \bibinfo{person}{Shohreh Deldari}, \bibinfo{person}{Linyao Chen}, \bibinfo{person}{Hao Xue}, {and} \bibinfo{person}{Flora~D. Salim}.} \bibinfo{year}{2025}\natexlab{a}.
\newblock \showarticletitle{{SensorLLM}: Aligning Large Language Models with Motion Sensors for Human Activity Recognition}. In \bibinfo{booktitle}{\emph{Proceedings of the 2025 Conference on Empirical Methods in Natural Language Processing}}. \bibinfo{publisher}{Association for Computational Linguistics}, \bibinfo{address}{Suzhou, China}, \bibinfo{pages}{354--379}.
\newblock
\href{https://doi.org/10.18653/v1/2025.emnlp-main.19}{doi:\nolinkurl{10.18653/v1/2025.emnlp-main.19}}


\bibitem[Lin(2004)]%
        {lin2004rouge}
\bibfield{author}{\bibinfo{person}{Chin-Yew Lin}.} \bibinfo{year}{2004}\natexlab{}.
\newblock \showarticletitle{{ROUGE}: A Package for Automatic Evaluation of Summaries}. In \bibinfo{booktitle}{\emph{Text Summarization Branches Out}}. \bibinfo{publisher}{Association for Computational Linguistics}, \bibinfo{address}{Barcelona, Spain}, \bibinfo{pages}{74--81}.
\newblock
\urldef\tempurl%
\url{https://aclanthology.org/W04-1013/}
\showURL{%
\tempurl}


\bibitem[Moon et~al\mbox{.}(2023)]%
        {moon2023imu2clip}
\bibfield{author}{\bibinfo{person}{Seungwhan Moon}, \bibinfo{person}{Andrea Madotto}, \bibinfo{person}{Zhaojiang Lin}, \bibinfo{person}{Aparajita Saraf}, \bibinfo{person}{Amy Bearman}, {and} \bibinfo{person}{Babak Damavandi}.} \bibinfo{year}{2023}\natexlab{}.
\newblock \showarticletitle{{IMU2CLIP}: Language-Grounded Motion Sensor Translation with Multimodal Contrastive Learning}. In \bibinfo{booktitle}{\emph{Findings of the Association for Computational Linguistics: EMNLP 2023}}. \bibinfo{publisher}{Association for Computational Linguistics}, \bibinfo{address}{Singapore}, \bibinfo{pages}{13246--13253}.
\newblock
\href{https://doi.org/10.18653/v1/2023.findings-emnlp.883}{doi:\nolinkurl{10.18653/v1/2023.findings-emnlp.883}}


\bibitem[Nie et~al\mbox{.}(2023)]%
        {nie2023patchtst}
\bibfield{author}{\bibinfo{person}{Yuqi Nie}, \bibinfo{person}{Nam~H. Nguyen}, \bibinfo{person}{Phanwadee Sinthong}, {and} \bibinfo{person}{Jayant Kalagnanam}.} \bibinfo{year}{2023}\natexlab{}.
\newblock \showarticletitle{A Time Series Is Worth 64 Words: Long-Term Forecasting with Transformers}. In \bibinfo{booktitle}{\emph{International Conference on Learning Representations}}.
\newblock
\urldef\tempurl%
\url{https://openreview.net/forum?id=Jbdc0vTOcol}
\showURL{%
\tempurl}


\bibitem[Ord{\'o}{\~n}ez and Roggen(2016)]%
        {ordonez2016deepconvlstm}
\bibfield{author}{\bibinfo{person}{Francisco~Javier Ord{\'o}{\~n}ez} {and} \bibinfo{person}{Daniel Roggen}.} \bibinfo{year}{2016}\natexlab{}.
\newblock \showarticletitle{Deep Convolutional and {LSTM} Recurrent Neural Networks for Multimodal Wearable Activity Recognition}.
\newblock \bibinfo{journal}{\emph{Sensors}} \bibinfo{volume}{16}, \bibinfo{number}{1} (\bibinfo{year}{2016}), \bibinfo{pages}{115}.
\newblock
\href{https://doi.org/10.3390/s16010115}{doi:\nolinkurl{10.3390/s16010115}}


\bibitem[Papineni et~al\mbox{.}(2002)]%
        {papineni2002bleu}
\bibfield{author}{\bibinfo{person}{Kishore Papineni}, \bibinfo{person}{Salim Roukos}, \bibinfo{person}{Todd Ward}, {and} \bibinfo{person}{Wei-Jing Zhu}.} \bibinfo{year}{2002}\natexlab{}.
\newblock \showarticletitle{{BLEU}: A Method for Automatic Evaluation of Machine Translation}. In \bibinfo{booktitle}{\emph{Proceedings of the 40th Annual Meeting of the Association for Computational Linguistics}}. \bibinfo{publisher}{Association for Computational Linguistics}, \bibinfo{address}{Philadelphia, Pennsylvania, USA}, \bibinfo{pages}{311--318}.
\newblock
\href{https://doi.org/10.3115/1073083.1073135}{doi:\nolinkurl{10.3115/1073083.1073135}}


\bibitem[Patsch et~al\mbox{.}(2025)]%
        {patsch2025wacu}
\bibfield{author}{\bibinfo{person}{Constantin Patsch}, \bibinfo{person}{Jaden Goter}, \bibinfo{person}{Joseph Greer}, \bibinfo{person}{Lingni Ma}, {and} \bibinfo{person}{Raj Sodhi}.} \bibinfo{year}{2025}\natexlab{}.
\newblock \showarticletitle{{WACU}: Multi-Modal Wristband Assistant for Contextual Understanding}. In \bibinfo{booktitle}{\emph{2025 IEEE/CVF International Conference on Computer Vision Workshops}}. \bibinfo{publisher}{IEEE}, \bibinfo{address}{Piscataway, NJ, USA}, \bibinfo{pages}{7273--7282}.
\newblock
\href{https://doi.org/10.1109/ICCVW69036.2025.00749}{doi:\nolinkurl{10.1109/ICCVW69036.2025.00749}}


\bibitem[Pu et~al\mbox{.}(2025)]%
        {pu2025promemassist}
\bibfield{author}{\bibinfo{person}{Kevin Pu}, \bibinfo{person}{Ting Zhang}, \bibinfo{person}{Naveen Sendhilnathan}, \bibinfo{person}{Sebastian Freitag}, \bibinfo{person}{Raj Sodhi}, {and} \bibinfo{person}{Tanya~R. Jonker}.} \bibinfo{year}{2025}\natexlab{}.
\newblock \showarticletitle{ProMemAssist: Exploring Timely Proactive Assistance Through Working Memory Modeling in Multi-Modal Wearable Devices}. In \bibinfo{booktitle}{\emph{Proceedings of the 38th Annual ACM Symposium on User Interface Software and Technology}} \emph{(\bibinfo{series}{UIST '25})}. \bibinfo{publisher}{Association for Computing Machinery}, Article \bibinfo{articleno}{56}, \bibinfo{numpages}{19}~pages.
\newblock
\href{https://doi.org/10.1145/3746059.3747770}{doi:\nolinkurl{10.1145/3746059.3747770}}


\bibitem[Reichman et~al\mbox{.}(2025)]%
        {reichman2025sensorqa}
\bibfield{author}{\bibinfo{person}{Benjamin Reichman}, \bibinfo{person}{Xiaofan Yu}, \bibinfo{person}{Lanxiang Hu}, \bibinfo{person}{Jack Truxal}, \bibinfo{person}{Atishay Jain}, \bibinfo{person}{Rushil Chandrupatla}, \bibinfo{person}{Tajana~{\v{S}}imuni{\'c} Rosing}, {and} \bibinfo{person}{Larry Heck}.} \bibinfo{year}{2025}\natexlab{}.
\newblock \showarticletitle{{SensorQA}: A Question Answering Benchmark for Daily-Life Monitoring}. In \bibinfo{booktitle}{\emph{Proceedings of the 23rd ACM Conference on Embedded Networked Sensor Systems}} \emph{(\bibinfo{series}{SenSys '25})}. \bibinfo{publisher}{Association for Computing Machinery}, \bibinfo{numpages}{8}~pages.
\newblock
\href{https://doi.org/10.1145/3715014.3722074}{doi:\nolinkurl{10.1145/3715014.3722074}}


\bibitem[Salter et~al\mbox{.}(2024)]%
        {salter2024emg2pose}
\bibfield{author}{\bibinfo{person}{Sasha Salter}, \bibinfo{person}{Richard Warren}, \bibinfo{person}{Collin Schlager}, \bibinfo{person}{Adrian Spurr}, \bibinfo{person}{Shangchen Han}, \bibinfo{person}{Rohin Bhasin}, \bibinfo{person}{Yujun Cai}, \bibinfo{person}{Peter Walkington}, \bibinfo{person}{Anuoluwapo Bolarinwa}, \bibinfo{person}{Robert Wang}, \bibinfo{person}{Nathan Danielson}, \bibinfo{person}{Josh Merel}, \bibinfo{person}{Eftychios Pnevmatikakis}, {and} \bibinfo{person}{Jesse Marshall}.} \bibinfo{year}{2024}\natexlab{}.
\newblock \showarticletitle{emg2pose: A Large and Diverse Benchmark for Surface Electromyographic Hand Pose Estimation}. In \bibinfo{booktitle}{\emph{Advances in Neural Information Processing Systems}}, Vol.~\bibinfo{volume}{37}. \bibinfo{publisher}{Curran Associates, Inc.}, \bibinfo{address}{Red Hook, NY, USA}, \bibinfo{numpages}{26}~pages.
\newblock
\urldef\tempurl%
\url{https://proceedings.neurips.cc/paper_files/paper/2024/hash/64f8884f6ba3d9ace5bb647c4f917896-Abstract-Datasets_and_Benchmarks_Track.html}
\showURL{%
\tempurl}


\bibitem[Tian et~al\mbox{.}(2025)]%
        {tian2025dailyllm}
\bibfield{author}{\bibinfo{person}{Ye Tian}, \bibinfo{person}{Xiaoyuan Ren}, \bibinfo{person}{Zihao Wang}, \bibinfo{person}{Onat Gungor}, \bibinfo{person}{Xiaofan Yu}, {and} \bibinfo{person}{Tajana Rosing}.} \bibinfo{year}{2025}\natexlab{}.
\newblock \showarticletitle{Daily{LLM}: Context-Aware Activity Log Generation Using Multi-Modal Sensors and {LLM}s}. In \bibinfo{booktitle}{\emph{2025 IEEE 22nd International Conference on Mobile Ad-Hoc and Smart Systems}}. \bibinfo{publisher}{IEEE}, \bibinfo{address}{Piscataway, NJ, USA}, \bibinfo{pages}{372--380}.
\newblock
\href{https://doi.org/10.1109/MASS66014.2025.00060}{doi:\nolinkurl{10.1109/MASS66014.2025.00060}}


\bibitem[Vaswani et~al\mbox{.}(2017)]%
        {vaswani2017attention}
\bibfield{author}{\bibinfo{person}{Ashish Vaswani}, \bibinfo{person}{Noam Shazeer}, \bibinfo{person}{Niki Parmar}, \bibinfo{person}{Jakob Uszkoreit}, \bibinfo{person}{Llion Jones}, \bibinfo{person}{Aidan~N. Gomez}, \bibinfo{person}{Lukasz Kaiser}, {and} \bibinfo{person}{Illia Polosukhin}.} \bibinfo{year}{2017}\natexlab{}.
\newblock \showarticletitle{Attention Is All You Need}. In \bibinfo{booktitle}{\emph{Advances in Neural Information Processing Systems}}, Vol.~\bibinfo{volume}{30}. \bibinfo{pages}{5998--6008}.
\newblock
\urldef\tempurl%
\url{https://proceedings.neurips.cc/paper/2017/hash/3f5ee243547dee91fbd053c1c4a845aa-Abstract.html}
\showURL{%
\tempurl}


\bibitem[Wang et~al\mbox{.}(2024)]%
        {wang2024ubiphysio}
\bibfield{author}{\bibinfo{person}{Chongyang Wang}, \bibinfo{person}{Yuan Feng}, \bibinfo{person}{Lingxiao Zhong}, \bibinfo{person}{Siyi Zhu}, \bibinfo{person}{Chi Zhang}, \bibinfo{person}{Siqi Zheng}, \bibinfo{person}{Chen Liang}, \bibinfo{person}{Yuntao Wang}, \bibinfo{person}{Chengqi He}, \bibinfo{person}{Chun Yu}, {and} \bibinfo{person}{Yuanchun Shi}.} \bibinfo{year}{2024}\natexlab{}.
\newblock \showarticletitle{{UbiPhysio}: Support Daily Functioning, Fitness, and Rehabilitation with Action Understanding and Feedback in Natural Language}.
\newblock \bibinfo{journal}{\emph{Proceedings of the ACM on Interactive, Mobile, Wearable and Ubiquitous Technologies}} \bibinfo{volume}{8}, \bibinfo{number}{1}, Article \bibinfo{articleno}{20} (\bibinfo{year}{2024}), \bibinfo{numpages}{27}~pages.
\newblock
\href{https://doi.org/10.1145/3643552}{doi:\nolinkurl{10.1145/3643552}}


\bibitem[Wang et~al\mbox{.}(2026)]%
        {wang2026llmsemg}
\bibfield{author}{\bibinfo{person}{Ming Wang}, \bibinfo{person}{Haoxuan Qu}, \bibinfo{person}{Qiuhong Ke}, \bibinfo{person}{Wei Zhou}, \bibinfo{person}{Hossein Rahmani}, {and} \bibinfo{person}{Jun Liu}.} \bibinfo{year}{2026}\natexlab{}.
\newblock \showarticletitle{Translating Signals to Languages for s{EMG}-Based Activity Recognition}. In \bibinfo{booktitle}{\emph{Proceedings of the IEEE/CVF Conference on Computer Vision and Pattern Recognition}}. \bibinfo{publisher}{IEEE}, \bibinfo{address}{Piscataway, NJ, USA}, \bibinfo{pages}{9317--9329}.
\newblock
\urldef\tempurl%
\url{https://openaccess.thecvf.com/content/CVPR2026/html/Wang_Translating_Signals_to_Languages_for_sEMG-Based_Activity_Recognition_CVPR_2026_paper.html}
\showURL{%
\tempurl}


\bibitem[Wang et~al\mbox{.}(2025)]%
        {wang2025reactemg}
\bibfield{author}{\bibinfo{person}{Runsheng Wang}, \bibinfo{person}{Xinyue Zhu}, \bibinfo{person}{Ava Chen}, \bibinfo{person}{Jingxi Xu}, \bibinfo{person}{Lauren Winterbottom}, \bibinfo{person}{Dawn~M. Nilsen}, \bibinfo{person}{Joel Stein}, {and} \bibinfo{person}{Matei Ciocarlie}.} \bibinfo{year}{2025}\natexlab{}.
\newblock \showarticletitle{{ReactEMG}: Zero-Shot, Low-Latency Intent Detection via sEMG}.
\newblock \bibinfo{journal}{\emph{arXiv preprint arXiv:2506.19815}} (\bibinfo{year}{2025}).
\newblock
\urldef\tempurl%
\url{https://arxiv.org/abs/2506.19815}
\showURL{%
\tempurl}


\bibitem[Yang et~al\mbox{.}(2025)]%
        {yang2025qwen25}
\bibfield{author}{\bibinfo{person}{An Yang}, \bibinfo{person}{Baosong Yang}, \bibinfo{person}{Beichen Zhang}, \bibinfo{person}{Binyuan Hui}, \bibinfo{person}{Bo Zheng}, \bibinfo{person}{Bowen Yu}, \bibinfo{person}{Chengyuan Li}, \bibinfo{person}{Dayiheng Liu}, \bibinfo{person}{Fei Huang}, \bibinfo{person}{Haoran Wei}, \bibinfo{person}{Huan Lin}, \bibinfo{person}{Jian Yang}, \bibinfo{person}{Jianhong Tu}, \bibinfo{person}{Jianwei Zhang}, \bibinfo{person}{Jianxin Yang}, \bibinfo{person}{Jiaxi Yang}, \bibinfo{person}{Jingren Zhou}, \bibinfo{person}{Junyang Lin}, \bibinfo{person}{Kai Dang}, \bibinfo{person}{Keming Lu}, \bibinfo{person}{Keqin Bao}, \bibinfo{person}{Kexin Yang}, \bibinfo{person}{Le Yu}, \bibinfo{person}{Mei Li}, \bibinfo{person}{Mingfeng Xue}, \bibinfo{person}{Pei Zhang}, \bibinfo{person}{Qin Zhu}, \bibinfo{person}{Rui Men}, \bibinfo{person}{Runji Lin}, \bibinfo{person}{Tianhao Li}, \bibinfo{person}{Tianyi Tang}, \bibinfo{person}{Tingyu Xia}, \bibinfo{person}{Xingzhang Ren},
  \bibinfo{person}{Xuancheng Ren}, \bibinfo{person}{Yang Fan}, \bibinfo{person}{Yang Su}, \bibinfo{person}{Yichang Zhang}, \bibinfo{person}{Yu Wan}, \bibinfo{person}{Yuqiong Liu}, \bibinfo{person}{Zeyu Cui}, \bibinfo{person}{Zhenru Zhang}, {and} \bibinfo{person}{Zihan Qiu}.} \bibinfo{year}{2025}\natexlab{}.
\newblock \showarticletitle{{Qwen2.5} Technical Report}.
\newblock \bibinfo{journal}{\emph{arXiv preprint arXiv:2412.15115}} (\bibinfo{year}{2025}).
\newblock
\urldef\tempurl%
\url{https://arxiv.org/abs/2412.15115}
\showURL{%
\tempurl}


\bibitem[Zhang et~al\mbox{.}(2026)]%
        {zhang2026fusionreview}
\bibfield{author}{\bibinfo{person}{Honglei Zhang}, \bibinfo{person}{Sidi~Mohamed Sid'El~Moctar}, \bibinfo{person}{Sofiane Boudaoud}, {and} \bibinfo{person}{Imad Rida}.} \bibinfo{year}{2026}\natexlab{}.
\newblock \showarticletitle{A Comprehensive Review of s{EMG}--{IMU} Sensor Fusion for Upper Limb Movements Pattern Recognition}.
\newblock \bibinfo{journal}{\emph{Information Fusion}}  \bibinfo{volume}{125}, Article \bibinfo{articleno}{103422} (\bibinfo{year}{2026}), \bibinfo{numpages}{43}~pages.
\newblock
\href{https://doi.org/10.1016/j.inffus.2025.103422}{doi:\nolinkurl{10.1016/j.inffus.2025.103422}}


\bibitem[Zhang et~al\mbox{.}(2020)]%
        {zhang2020bertscore}
\bibfield{author}{\bibinfo{person}{Tianyi Zhang}, \bibinfo{person}{Varsha Kishore}, \bibinfo{person}{Felix Wu}, \bibinfo{person}{Kilian~Q. Weinberger}, {and} \bibinfo{person}{Yoav Artzi}.} \bibinfo{year}{2020}\natexlab{}.
\newblock \showarticletitle{{BERTScore}: Evaluating Text Generation with {BERT}}. In \bibinfo{booktitle}{\emph{International Conference on Learning Representations}}.
\newblock
\urldef\tempurl%
\url{https://openreview.net/forum?id=SkeHuCVFDr}
\showURL{%
\tempurl}


\bibitem[Zhang et~al\mbox{.}(2025)]%
        {zhang2025sensorlm}
\bibfield{author}{\bibinfo{person}{Yuwei Zhang}, \bibinfo{person}{Kumar Ayush}, \bibinfo{person}{Siyuan Qiao}, \bibinfo{person}{A.~Ali Heydari}, \bibinfo{person}{Girish Narayanswamy}, \bibinfo{person}{Max Xu}, \bibinfo{person}{Ahmed Metwally}, \bibinfo{person}{Jinhua Xu}, \bibinfo{person}{Jake Garrison}, \bibinfo{person}{Xuhai~{``Orson''} Xu}, \bibinfo{person}{Tim Althoff}, \bibinfo{person}{Yun Liu}, \bibinfo{person}{Pushmeet Kohli}, \bibinfo{person}{Jiening Zhan}, \bibinfo{person}{Mark Malhotra}, \bibinfo{person}{Shwetak Patel}, \bibinfo{person}{Cecilia Mascolo}, \bibinfo{person}{Xin Liu}, \bibinfo{person}{Daniel McDuff}, {and} \bibinfo{person}{Yuzhe Yang}.} \bibinfo{year}{2025}\natexlab{}.
\newblock \showarticletitle{{SensorLM}: Learning the Language of Wearable Sensors}. In \bibinfo{booktitle}{\emph{Advances in Neural Information Processing Systems}}, Vol.~\bibinfo{volume}{38}.
\newblock
\urldef\tempurl%
\url{https://proceedings.neurips.cc/paper_files/paper/2025/hash/42cd98f0e7520d4a63c34891ac1c972f-Abstract-Conference.html}
\showURL{%
\tempurl}


\end{thebibliography}

\end{document}